%% file: root_copyright.tex
\documentclass[conference]{IEEEtran}
\IEEEoverridecommandlockouts
\usepackage{cite}
\usepackage{amsmath,amssymb,amsfonts}
\usepackage{algorithmic}
\usepackage{graphicx}
\usepackage{textcomp}
\usepackage{xcolor}
\usepackage{booktabs}
\usepackage{svg}
\usepackage{subcaption}
\usepackage{graphicx}  
\usepackage{xcolor}
\usepackage{multirow}
\usepackage{array}
\usepackage{changes}
\newcommand\copyrighttext{
	\footnotesize \textcopyright 2025 IEEE. Personal use of this material is permitted.  Permission from IEEE must be obtained for all other uses, in any current or future media, including reprinting/republishing this material for advertising or promotional purposes, creating new collective works, for resale or redistribution to servers or lists, or reuse of any copyrighted component of this work in other works.
	DOI: 10.1109/IV64158.2025.11097683 }
\newcommand\copyrightnotice{
	\begin{tikzpicture}[remember picture,overlay]
		\node[anchor=south,yshift=10pt] at (current page.south) {\fbox{\parbox{\dimexpr\textwidth-\fboxsep-\fboxrule\relax}{\copyrighttext}}};
	\end{tikzpicture}
}

\def\BibTeX{{\rm B\kern-.05em{\sc i\kern-.025em b}\kern-.08em
    T\kern-.1667em\lower.7ex\hbox{E}\kern-.125emX}}
   
\begin{document}
 
\title{Reliability comparison of vessel trajectory prediction models via Probability of Detection
}
\author{\IEEEauthorblockN{Zahra Rastin}
\IEEEauthorblockA{\textit{Chair of Dynamics and Control} \\
\textit{University of Duisburg-Essen}\\
Duisburg, Germany \\
zahra.rastin@uni-due.de}
\and
\IEEEauthorblockN{Kathrin Donandt}
\IEEEauthorblockA{\textit{Inst. for Sustainable and Autonomous Maritime Systems} \\
\textit{University of Duisburg-Essen}\\
Duisburg, Germany \\
kathrin.donandt@uni-due.de}
\and
\IEEEauthorblockN{Dirk Söffker}
\IEEEauthorblockA{\textit{Chair of Dynamics and Control} \\
\textit{University of Duisburg-Essen}\\
Duisburg, Germany \\
soeffker@uni-due.de}
}

\maketitle
\copyrightnotice
\begin{abstract}
This contribution addresses vessel trajectory prediction (VTP), focusing on the evaluation of different deep learning-based approaches. The objective is to assess model performance in diverse traffic complexities and compare the reliability of the approaches. While previous VTP models overlook the specific traffic situation complexity and lack reliability assessments, this research uses a probability of detection analysis to quantify model reliability in varying traffic scenarios, thus going beyond common error distribution analyses. All models are evaluated on test samples categorized according to their traffic situation during the prediction horizon, with performance metrics and reliability estimates obtained for each category. The results of this comprehensive evaluation provide a deeper understanding of the strengths and weaknesses of the different prediction approaches, along with their reliability in terms of the prediction horizon lengths for which safe forecasts can be guaranteed. These findings can inform the development of more reliable vessel trajectory prediction approaches, enhancing safety and efficiency in future inland waterways navigation.
\end{abstract}


\input{sections/introduction_kathrin_final_v0}
\input{sections/related_work_kathrin_final_v0}

\input{sections/method}
\input{sections/results}
\input{sections/conclusion_soeffker}
\input{sections/acknowledgement}
\bibliographystyle{IEEEtran} 
\bibliography{IEEEabrv,bibliography}

\end{document}

%% file: sections/introduction_kathrin_final_v0.tex
\section{INTRODUCTION}
The ability to predict vessel movements around the own ship in complex traffic environments is crucial for effective decision making to maintain and enhance traffic safety. In recent years, many deep learning-based approaches have been proposed for vessel trajectory prediction (VTP), with increasing attention being paid to the behavior of interacting self-acting systems \cite{Feng.2022,Liu.2024,Zhang.2024,Jiang2024,Liu.2023,Wang.2023}. 
River navigation poses particular challenges for interacting vessels due to limited space, river currents and site-specific rules for encounters. In order to gain a comprehensive understanding of the performance of an inland VTP model, an assessment in relation to the complexity of traffic situations is crucial. 
A common limitation of VTP approaches is, however,  that quantitative assessment focuses on mean displacement errors \cite{Zhang.2022}, which is calculated over the entire test dataset without consideration of the traffic complexity. 
To gain a deeper insight into the reliability of the model predictions, a probability of detection (POD) approach can be applied. The POD method is a statistical approach to evaluate the performance and reliability of diagnostic systems. It quantifies the probability that a detection/(prediction) system will correctly identify a target, taking into account the variability and uncertainty of the process. Originally developed to evaluate non-destructive testing (NDT) methods, the POD approach has been adopted in safety-critical areas such as aerospace and defense \cite{doi:10.1177/14759217231193088                                                      }, \cite{hdbk2009nondestructive}. It is receiving increasing attention in areas such as structural health monitoring and the nuclear industry \cite{doi:10.1177/14759217211060780                                                       }, \cite{s23104813} and has also been used to evaluate and compare ML classifiers \cite{9791231}, \cite{automation5030019}. In this study, the evaluation limitations in the previous VTP approaches are addressed with a focus on the inland shipping. Performance is compared in different traffic situations by error statistics, including box and scatter plots, and using a POD approach. 

%% file: sections/related_work_kathrin_final_v0.tex
\section{RELATED WORK}
\label{sec:related_work}

Research on inland VTP increasingly uses deep learning, mainly modeling individual vessel trajectories based on historical AIS data. In \cite{You.2020}, LSTM- and GRU-based encoder-decoder models for predicting vessel trajectories are compared. In \cite{Dijt.2020}, a CNN-LSTM combination for inland VTP considering radar and IENC data alongside AIS data is employed. In \cite{Donandt.2023ITSC}, a transformer encoder-decoder model processes vessel positions and speeds transformed to inland-specific reference systems. Distribution curves representing inland vessel behavior are additionally processed by an RNN of which the hidden state is used as further input to a transformer encoder-decoder model for VTP in \cite{Donandt.2024}. Interaction modeling in inland VTP remains rare. Interaction-aware VTP models developed for river and seagoing vessels include \cite{Feng.2022} and \cite{Jiang2024}, where graph neural networks and temporal CNNs are combined. In \cite{Donandt.2023SMC}, the model proposed in \cite{Donandt.2023ITSC} is extended by considering  surrounding traffic through a social tensor transformer. 
Even if some of the mainly maritime-oriented interaction-aware VTP studies
illustrate performance in specific traffic scenarios \cite{ Feng.2022,Zhang.2024,Jia.2023,Liu.2024}, 
the selection strategy of these scenarios is often unspecified or arbitrary and the corresponding results may not reflect the predictive accuracy of the models representatively enough. A further but related problem with undifferentiated assessment is that the error distribution is often not specified. Furthermore, some evaluations rely only on the best prediction from a set of prediction candidates \cite{Ma.2024, Sigillo.2023}, which can give an idealized view of the model's performance. 
 
POD-based evaluation of ML models is conducted in \cite{9791231} where the POD approach is applied to assess and compare the performance of various classifiers and their improved versions, in predicting human driver lane changing behavior, in terms of how early they can reliably predict the correct behavior. In \cite{automation5030019}, a method aimed at further enhancing the performance of the ML models studied in \cite{9791231} is proposed. 
The POD approach is 
used to evaluate the proposed method and compare it to the improved ML-based lane changing behavior prediction method investigated in the aforementioned work. Here, the POD approach is used to compare the reliability of ML-based VTP algorithms based on the maximum prediction horizon they
can achieve without exceeding a predefined error threshold. This simplifies the performance evaluation to a single, easy-to-interpret metric.

%% file: sections/method.tex
\section{METHODS}\label{sec:methods}
\subsection{Vessel trajectory prediction models}
Vessel trajectory prediction can be defined as a time series forecasting task in which an input time series of length $T_{in}$ is used to generate an output time series of length $T_{out}$. Both consist of a sequence of features describing a vessel's navigation status at a given time instant. These features can differ in the input and output sequence. In its simplest form, the navigation status features are the vessel's longitudinal and lateral position. 
The VTP models compared in this study, STT-R-CSCT \cite{Donandt.2023SMC}, N-CSCT \cite{Donandt.2023ITSC}, and GMM-Trans-GRU \cite{Donandt.2024}, differ in the navigation status features used and specific sub-modules introduced into the model architecture to process specific navigation status features.

\subsubsection{Navigation status definitions}

\begin{table*}[h!]
\caption{Navigation status features and additional context information per model }
\label{tab:features}
\begin{center}
\renewcommand{\arraystretch}{1.5} 
\begin{tabular}{p{3.25cm}p{4.125cm}p{4.125cm}p{5cm}}\toprule & \textbf{N-CSCT \cite{Donandt.2023ITSC}}                                                   & \textbf{STT-R-CSCT \cite{Donandt.2023SMC}} & \textbf{GMM-Trans-GRU \cite{Donandt.2024}}                                                                                                                             \\ \midrule
\textbf{Longitudinal dislocation} & Difference between typical and actual WW-km distance per minute                          & WW-km distance per minute                                                                                & WW-km distance per minute                                                                                                                                               \\
\textbf{Lateral (dis-)location}      & Change rate of distance from typical route                                               & Change rate of relative distance from fairway boundary                                                   & Fairway center offset                                                                                                                                                   \\
\textbf{Additional features}           & Euclidean distance between subsequent positions on typical route with 0.1 WW-km distance & \textit{Social tensor}                                               & River curvature and curve orientation; Vectors representing distribution of fairway center offset and WW-km distance per minute \\
\textbf{Context information}     & River curvature and curve orientation vector                                             & River curvature and curve orientation vector & None \\
\bottomrule
\end{tabular}
\end{center}
\end{table*}
Defining the vessel location in terms of a global reference such as the Universal Transverse Mercator (UTM) coordinate system poses problems especially for inland VTP models. The navigation space is constrained by river boundaries and the river orientation changes frequently. With a global location information, the model would be unable to detect invalid lateral deviations due to river boundaries and similarities in trajectories located in river sections with different orientations (e.g. North-South, East-West). Therefore, a river- and navigation-specific reference system for vessel position definition is introduced in \cite{Donandt.2023ITSC}. The river- and navigation-specific context-sensitive classification transformer (R-CSCT and N-CSCT) make use of these reference systems and are based on the transformer introduced in \cite{10.5555/3295222.3295349   } for sequence modeling tasks as an alternative to recurrent or convolutional neural networks-based encoder-decoder models. Whereas the R-CSCT uses the waterway kilometer (WW-km) distance and change rate of the fairway boundary offset between subsequent time steps as navigation status features, the N-CSCT uses relative to a typical route and typical velocity, respectively, which are generated through a statistical evaluation of AIS data. The GMM-Trans-GRU model uses similar (dis-)location featurs as the R-CSCT, replacing only the fairway boundary offset change rate by the fairway center offset.  
Apart from these (dis-)location features, the navigation status of each model contains additional features. The N-CSCT includes the Euclidean distance between subsequent positions of a distance of 0.1 WW-km on the typical route - necessary to mitigate issues related to the non-equidistance of subsequent waterway hectometers in curves. 
The STT-R-CSCT model, which is an extended R-CSCT able to accounts for the traffic situation, utilizes a \textit{social tensor} \cite{Alahi2016} as additional input per time step.  
The GMM-Trans-GRU model uses vectors derived from density curves of the fairway center offset and WW-km distance per minute. These density curves are obtained through Gaussian Mixture Models (GMM) fitted on AIS data for specific discharge levels and each waterway hectometer.  
All models incorporate river curvature and curve orientation values, either as further navigation status features (GMM-Trans-GRU), or as context information used for the transformer decoder initialization (STT-R-CSCT and N-CSCT). 
Details on the different navigation status features are given in \cite{Donandt.2023ITSC}, \cite{Donandt.2024}, and \cite{Donandt.2023SMC}. In Table \ref{tab:features}, the navigation status features used by each model are summarized. 

\subsubsection{Model architectures}
\begin{figure}[!h]
    \begin{subfigure}[t]{0.5\textwidth}
    \includegraphics[trim=0.7cm 0cm 0.7cm 0.67cm, clip]
    {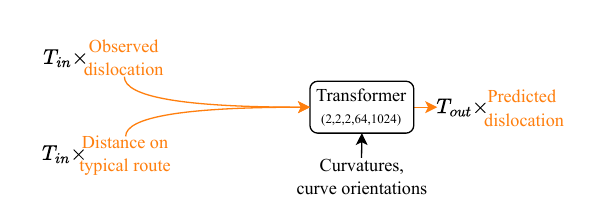}
    \caption{N-CSCT}\end{subfigure}
    \vspace{10pt}
    \begin{subfigure}[t]{0.5\textwidth}
    \centering 
    \includegraphics[trim=0.7cm 0cm 0.5cm 0.67cm, clip]{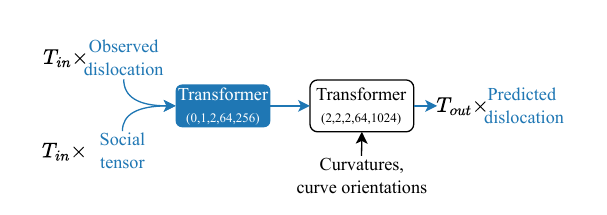} 
    \caption{STT-R-CSCT}\label{fig:stt_r_csct}
    \end{subfigure}%
    \vspace{10pt}
    \begin{subfigure}[t]{0.5\textwidth}
\includegraphics[trim=0.7cm 0cm 0.7cm 0.67cm, clip]{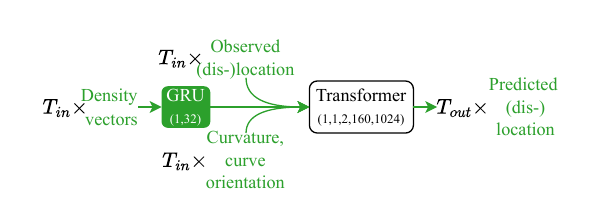} 
\caption{GMM-Trans-GRU}\label{fig:gmm_trans_gru} 
\end{subfigure}
\caption{The compared VTP models. Numbers are encoder layers, decoder layers, attention heads, query/key/value dimension and feedforward neural network output shape (transformer), and layers and hidden state dimension (GRU).}\label{fig:model_archs}
\end{figure}

A schematic overview of the differences in the model architectures is shown in Fig. \ref{fig:model_archs}. The main component of all models is a transformer that learns the mapping of the input to the output time series. The STT-R-CSCT includes an additional transformer-based sub-module, the social tensor transformer (STT), which fuses the \textit{social tensor} with the vessel's dislocation features. The GMM-Trans-GRU processes 
the sequence of GMM-based density vectors for the river section the vessel is heading to with a GRU, and its final hidden state is fed, along with the vessel (dis-)location features, into the main transformer. Note, that different from STT-R-CSCT and N-CSCT, GMM-Trans-GRU has only one encoder and one decoder layer in its transformer. Increasing the number of layers is not resulting in an improved performance in this case. Also, the query/key/valud dimension is higher due to additional input (GRU hidden state).

\subsection{POD-based evaluation process}\label{sec:pod}

The performance of the VTP models is evaluated using standard statistical metrics and the POD approach which incorporates process parameters, task-specific factors affecting outcomes, distinct from training or model-specific hyperparameters. For instance, the prediction horizon, which affects forecast accuracy over time, is a key process parameter in VTP. 
The POD approach yields a curve called the POD curve, which in NDT shows the likelihood of detecting a flaw based on its size. This curve can be generated using either binary (hit/miss method) or continuous data (\^a versus a method).

\begin{figure*}[!h]
     \centering
     \begin{subfigure}[b]{0.3291\textwidth}
         \centering
         \includegraphics[width=\textwidth]{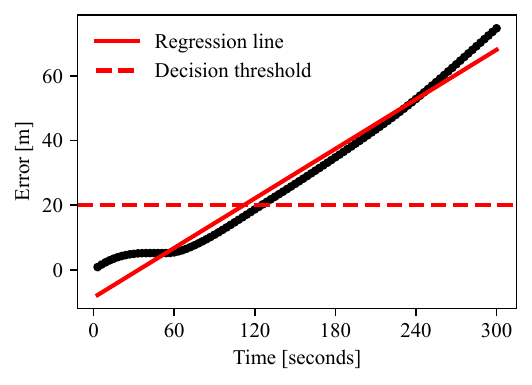}
         \caption{}
         \label{fig:regression line}
     \end{subfigure}
          \begin{subfigure}[b]{0.3291\textwidth}
         \centering
         \includegraphics[width=\textwidth]{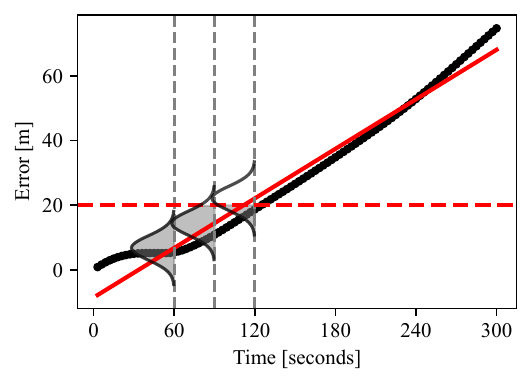}
         \caption{}
         \label{fig:pdfs}
     \end{subfigure}
     \begin{subfigure}[b]{0.3291\textwidth}
         \centering
         \includegraphics[width=\textwidth]{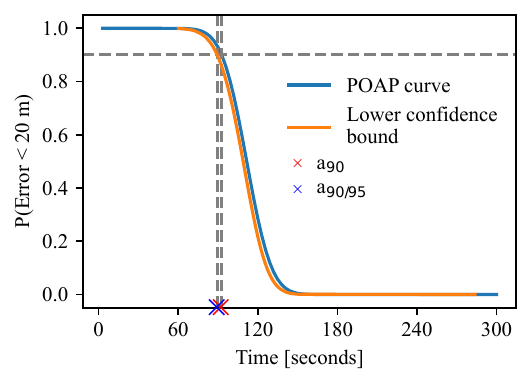}
         \caption{}
         \label{fig:pop curve}
     \end{subfigure}
    \caption{POD-based evaluation steps: (a) fitting a regression line to the data, (b) obtaining the probability of error below decision threshold at each time step with 50\% confidence, and (c) generating the POAP curve and its 95\% lower confidence bound.}
    \label{fig:pod process}
\end{figure*}

Considering the prediction horizon as the process parameter and the associated vessel trajectory prediction error as the response value, the â versus a approach to POD is taken to determine the maximum prediction horizon where the error remains reliably below a specified threshold. The key stages of the POD-based evaluation process for this purpose are illustrated in Fig. \ref{fig:pod process}. 
The first step involves plotting the prediction errors, $\hat{a}$, against the prediction horizons, $a$. Four combinations of logarithmic and Cartesian axes for $\hat{a}$ and $a$ are considered. The best solution that meets the criteria i) the data should lie close to a straight line and ii) the variance should be evenly distributed around the regression line is chosen.

Regression analysis is performed on the data from the selected graph using the maximum likelihood method (Fig. \ref{fig:regression line}). For the \^a vs. a scenario, the data can be represented as

\begin{equation}
\hat{a} = b + ma + \epsilon,
\label{eq:regression}
\end{equation}

where $b$ and $m$ are the regression coefficients, and $\epsilon\sim N(0,\tau)$ denotes the error term, which follows a normal distribution with a mean of 0 and a standard deviation of $\tau$. To assess the probability of reliable VTP at each horizon, the likelihood that the prediction error at that horizon stays below a specified decision threshold ($\hat{a}_{th}$) must be calculated. With 50\% confidence, this probability corresponds to the area under the probability density function (PDF), derived from the distribution of the error term in Eq. \ref{eq:regression} and centered around the regression line, that falls below $\hat{a}_{th}$ (
shaded region in Fig. \ref{fig:pdfs}). This probability at each prediction horizon is calculated as

\begin{equation}
\mathrm{P}(\hat{a} < \hat{a}_{th}) = \phi(\frac{\hat{a}_{th} - (b + ma)}{\tau}),
\label{eq:pop}
\end{equation}

where $\phi$ is the cumulative standard normal distribution function. As the prediction horizon increases, the probability obtained from Eq. \ref{eq:pop} decreases. Thus, unlike standard ascending POD curves, the curve obtained by plotting this probability against the prediction horizon will be descending. Hereafter, this curve is referred to as the probability of accurate prediction (POAP) curve, with the vertical axis representing the probability of predicting the vessel's trajectory with an error less than $\hat{a}_{th}$. 

To account for uncertainties in estimating the parameters of the regression line, the 95 percentile POAP curve is generated using the Wald method. This curve defines the boundary below which 95\% of the average POAP curves would fall if the study was repeated many times. A POAP curve alongside its 95\% lower confidence bound is depicted in Fig. \ref{fig:pop curve}. In this figure, $a_{90}$ and $a_{90/95}$ mark the prediction horizons at which the POAP curve and its 95\% lower confidence bound, respectively, achieve a 90\% probability. The $a_{90/95}$ values obtained for the VTP models under various traffic conditions are utilized to assess and compare their performance and reliability in this study.

%% file: sections/results.tex
\section{APPLICATION AND RESULTS}
\subsection{Dataset description}
An AIS dataset spanning from 01/01/2021 to 04/30/2024 and covering a section of the Rhine river (Rhine-km 595.5 to 613.5) is used. Trajectories are extracted by splitting at breaks and turns and subsequently interpolated to obtain homogeneous time steps of 1 min between subsequent trajectory positions. 
Only the trajectory prediction for the upstream direction is considered in this study. 
The trajectory of each upstream navigating vessel is fused with trajectories of all other vessels navigating at the same time in the considered river section. These traffic-scene enriched trajectories of 'ego' vessels are then split into sequences covering 10 time steps. 
Originally, $T_{in} = T_{out} = 5$ for each model. Due to the different lateral navigation status features used by the models (change rates for the CSCTs and absolute value in GMM-Trans-GRU, see Table \ref{tab:features}), the number of actually covered positions by 10 time steps differ. Therefore, for both CSCT variants, the first predicted WW-km is replaced by the ground truth and for the GMM-Trans-GRU, $T_{in}$ is set to 6, discarding the WW-km of the 5th prediction. 
All models are trained using 22k of the obtained sequences, and validated on 2.8k sequences. The 2.6k sequences of the test set are categorized by their traffic situation during prediction. 
A traffic situation refers here to a combination of ship interactions (encounter, overtaking, overtaken) from the perspective of the target vessel. For example, 'encounter-1 overtaken-1' means the target vessel passes a downstream vessels and is overtaken by an upstream vessel, with the order of interactions not being considered. In the following sections, the overall and traffic situation-specific prediction performance will be evaluated. 

\subsection{Standard statistical evaluation of VTP models}\label{sec:b}
\begin{figure}[h!]
\centering
    \includegraphics[width=0.4\textwidth, trim=0.4cm 0.4cm 0.3cm 0.35cm, clip]{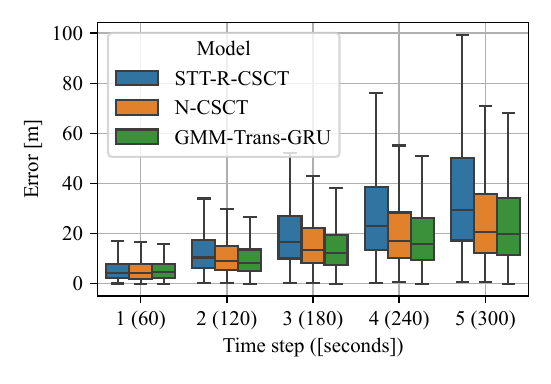}
    \caption{Overall distribution of the displacement error obtained per model over all prediction steps.}
    \label{fig:ade_boxplots} 
\end{figure}

First, the displacement errors of the VTP models, i.e. the Euclidean distance between the predicted and ground truth positions, are compared. 
According to Fig. \ref{fig:ade_boxplots}, the GMM-Trans-GRU model shows the lowest medians and lowest spread (indicated by the box plot whiskers) for prediction horizons over 1 min, while the interaction-aware (STT-R-CSCT) surprisingly performs the worst. 
In Table \ref{tab:interaction_case_errors}, the overall mean, median, and standard deviation of the displacement error at the last prediction step are compared to the ones obtained for situations where the target vessel only encounters downstream navigating vessels ('Encounter'), overtakes one or more vessels with optional encounters ('Overtaking'), and is overtaken by one or more vessels with optional encounters ('Overtaken'). All models struggle to predict well in the underrepresented overtaking and overtaken cases. The overall error statistics are thus creating a too optimistic impression of the models' capabilities. 
\begin{table}[]
\caption{Error statistics (mean, (median, std.) [m]) per model and traffic situation (number of samples in parentheses) at 5th prediction step - best results marked with *.}
\label{tab:interaction_case_errors}
\begin{center}
\begin{tabular}{p{1.4cm}p{1.7cm}p{1.7cm}p{2.3cm}}
\toprule
\parbox{1.3cm}{\textbf{Traffic\\situation}} & \textbf{STT-R-CSCT}                                             & \textbf{N-CSCT}                                                 & \textbf{GMM-Trans-GRU}                                         \\[0.1cm] \midrule
\parbox{1.3cm}{Encounter (2320)} & \parbox{1.7cm}{37.96 \\ (27.7, 35.01)} & \parbox{1.7cm}{27.58 \\ (19.63, 28.29)}  & \parbox{2.3cm}{25.05*\\ (18.72*, 22.79*)} \\[0.2cm]
\parbox{1.3cm}{Overtaking (149)}& \parbox{1.7cm}{57.63 \\ (44.81, 44.22)} & \parbox{1.7cm}{45.51 \\ (36.03, 36.05)} & \parbox{2.3cm}{42.55*\\ (32.14*, 31.17*)} \\[0.2cm]
\parbox{1.3cm}{Overtaken (109)} & \parbox{1.7cm}{68.07\\ (56.1, 45.85*)} & \parbox{1.7cm}{72.46\\ (58.57, 52.39)}  & \parbox{2.3cm}{64.87*\\ (55.21*, 50.38)} \\[0.1cm]
\midrule
\parbox{1.3cm}{Overall (2578)}  & \parbox{1.7cm}{40.37 \\ (29.44, 36.85)} & \parbox{1.7cm}{30.51 \\ (20.72, 31.7)}  & \parbox{2.3cm}{27.74*\\ (19.71*, 26.59*) }  \\ \bottomrule
\end{tabular}
\end{center}
\end{table}

The previous evaluations require careful interpretation of data distribution. The POD approach 
offers a more concise, straightforward method, providing deeper insights into the reliability of different VTP approaches.

\subsection{POD-based evaluation of VTP models}
\begin{figure}[htbp]
\centering
        \includegraphics[width=0.4\textwidth,trim=0.2cm 0.2cm 0.2cm 0.25cm, clip]{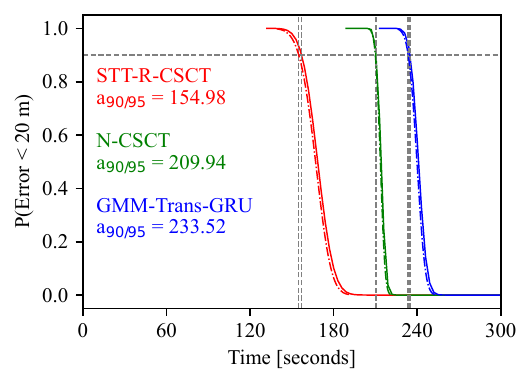}
    \caption{POAP curves (solid lines) with lower confidence bounds (dash-dot lines) under 'encounter-1' traffic situation. }
    \label{fig:pod curve examples}
\end{figure}

The POD approach (Sec. \ref{sec:pod}) is used for a concise evaluation and comparison of the VTP models. 
The predicted positions are first interpolated to generate a more  continuous trajectory with a 3-second sampling rate. 
For each VTP model and traffic scenario, prediction errors across all samples are calculated for each prediction horizon. These errors serve as the response values to generate the 
POAP curve. However, due to their scattered and diverse nature, the response values are averaged at each process parameter level (prediction horizon) to facilitate the fitting of the regression line, as a step in generating the POAP curve. The resulting data points can be effectively represented by a regression line. The decision threshold is set to 20 m, chosen as an example to determine how far the prediction horizon can extend while keeping the error reliably below the overall median of the best-performing model (GMM-Trans-GRU) at the final prediction time step considered (about 20 m). However, this threshold can and should be adjusted to any value of interest, depending on the specific context or requirements. 
Examples of POAP curves obtained for 'encounter-1' traffic scenario are shown in Fig. \ref{fig:pod curve examples}. In this figure, $a_{90/95}$ values represent prediction horizons until where prediction errors are safely below 20 m. A higher $a_{90/95}$ clearly indicates a better performance of the model, as it reflects a wider range over which the model remains reliable. 

The $a_{90/95}$ values from the VTP models under various traffic scenarios are presented in Table \ref{tab:pod results}. The best and worst $a_{90/95}$ values are marked with * and $^\dagger$ respectively. The table shows that in almost all cases, GMM-Trans-GRU model performs the best, while STT-R-CSCT model demonstrates the worst performance. This result is consistent with findings from the boxplots in Section \ref{sec:b} (Fig. \ref{fig:ade_boxplots}). 
Considering the overall performance, it can be seen from Table \ref{tab:pod results} that STT-R-CSCT, N-CSCT, and GMM-Trans-GRU models can reliably predict the vessel's trajectory up to prediction horizons of about 2.5, 3, and 3.5 min, while maintaining errors below 20 m. Although averaging response values at each process parameter level for generating POAP curves may slightly overestimate VTP model performance by ignoring response data scatter, this approach remains highly effective for comparing model performance concisely and selecting the best model for the specified task.

The POAP curves simplify the performance into a single, interpretable metric—such as the $a_{90/95}$ value. This makes the evaluation process more straightforward, allowing for objective comparison of approaches/models without requiring detailed analysis of box- and scatter plots. Additionally, the POD approach provides a means to specify a prediction horizon for VTP models, ensuring that the prediction error remains safely below a prespecified threshold. These features enable an efficient decision-making process, particularly in scenarios where quick and clear judgment is critical.
\begin{table}[]
\caption{Probability of detection-based performance evaluation results for VTP models under different traffic situations (number of samples in parentheses).}
\label{tab:pod results}
\begin{tabular}{p{2.2cm}p{1.65cm}p{1.05cm}p{2.275cm}}
\toprule
\multirow{2}{*}{\textbf{\shortstack{Traffic\\situation}}}& \multicolumn{3}{c}{\textbf{$a_{90/95}$ [min]}}
\\ 
\cline{2-4}
& 
\textbf{STT-R-CSCT}\rule{0pt}{2.5ex}&
\textbf{N-CSCT}&
\textbf{GMM-Trans-GRU}
\\
\cline{1-4}
Encounter-1 (1938)\rule{0pt}{2.5ex}&
2.583$^\dagger$&
3.499&
3.892*
\\
Overtaking-1 (106)&
1.742$^\dagger$&
2.07&
2.149*
\\
Overtaken-1 (78)&
1.53$^\dagger$&
1.541&
1.727*
\\
Encounter-2 (355)&
2.707$^\dagger$&
3.583&
4.134*
\\
Encounter-3 (24)&
3.116$^\dagger$&
4.767&
$>$ 5*
\\
Encounter-1 overtaking-1 (34)&
1.472$^\dagger$&
1.715&
1.73*
\\
Encounter-1 overtaken-1 (28)&
1.49*&
1.422&
1.375$^\dagger$
\\
\cline{1-4}
Overall (2578)\rule{0pt}{2.5ex}&
2.452$^\dagger$&
3.19&
3.541*
\\
\bottomrule
\end{tabular}
\end{table}

\subsection{Discussion}

The overall performance comparison reveals that the interaction-aware model exhibits weaknesses, as evidenced by its overall errors (Fig. \ref{fig:ade_boxplots} and Table \ref{tab:interaction_case_errors}), compared to models without social but more advanced spatial situation awareness (N-CSCT and GMM-Trans-GRU). For the prediction problem and scope considered here, providing the prediction models with information about typical navigation behavior in terms of typical routes/speeds or probability density curves of vessel locations on the river thus proves to be more beneficial than information about the surrounding vessels’ behavior. 
The performance of the models varies significantly depending on the traffic situation. The distribution of traffic scenarios is highly imbalanced, with encounter-only cases constituting 90\% of the test dataset (Table \ref{tab:interaction_case_errors}). Since the train-validation-test split is created randomly, the distribution in the training dataset is expected to be similar. This imbalance suggests that the models have difficulty learning the behavior patterns of the remaining 10\%. As a result, the overall evaluation of model performance has limited expressiveness. In addition to the dataset imbalance, overtaking or ceding passage to another vessel requires maneuvering. The lower performance in overtaken cases compared to the overtaking cases may stem from the increased uncertainty involved: A vessel being overtaken does not necessarily need to maneuver if enough space is available for the overtaking vessel to pass. This assumption is confirmed by the observed high standard deviations for this case (Table \ref{tab:interaction_case_errors}). 
The advantage of the interaction-aware model in overtaking and overtaken scenarios, where the behavior of surrounding vessels could provide valuable information for prediction, is not immediately evident. The STT-R-CSCT model performs worst or second worst in these cases. However, the performance decline from overtaking to overtaken scenarios is much smaller compared to the interaction-agnostic models (10.49 compared to 26.95 and 22.32 (mean), and the standard deviation is similar for both cases (Table \ref{tab:interaction_case_errors}). In contrast, the standard deviations increase significantly for N-CSCT and GMM-Trans-GRU. This suggests that considering interactions might offer a benefit in these cases. Adapting the STT-R-CSCT model to include enhanced spatial situational awareness could potentially improve its performance. Additionally, the prediction performance of all VTP models could be further improved through traffic situation-specific tuning, particularly for more challenging interaction scenarios. 

%% file: sections/conclusion_soeffker.tex
\section{CONCLUSION AND OUTLOOK}
This study represents an initial effort to evaluate the performance of inland VTP models across traffic situations of varying complexities. For the first time, the POD approach is used alongside error statistics to offer new insights into model reliability. A comparison of the performance evaluation results based on error statistics and POD shows that the VTP approach, which is interaction-agnostic (GMM-TransGRU), is recognized as the best performer in almost all traffic situations. The POD-based evaluation used provides the static parameter $a_{90/95}$ as a performance indicator for a simple model comparison. The determination of a time horizon within which the prediction error remains reliably below a specified threshold, $\hat{a}_{th}$, allows a comparison of the performance of the models. 
According to the POD evaluation, the models predict ship trajectories with an error of safely less than 20 m for periods ranging from 1.37 to more than 5 min, depending on the complexity of the traffic situation. 
The comparison of the VTP models shows that further improvements are needed, especially with regard to underrepresented, complex interaction scenarios. Future work needs to address this issue by improving the realization of interaction awareness and solving the unbalanced representation of different traffic situations in the dataset. The POD-based evaluation can be extended to consider further process parameters to evaluate different aspects of the performance of VTP models.

%% file: sections/acknowledgement.tex
\section*{ACKNOWLEDGMENT}
The authors thank the Federal Waterway Engineering and Research Institute (Germany) for providing AIS and river geometry data and the training infrastructure, and the German Federal Waterways and Shipping Agency and Federal Institute of Hydrology for the discharge measurement data.